%% file: main.tex
\documentclass[conference]{IEEEtran}
\IEEEoverridecommandlockouts

\input{include/packages}
\usepackage{cleveref}
\usepackage{enumitem}
\crefformat{section}{\S#2#1#3} 
\crefformat{subsection}{\S#2#1#3}
\crefformat{subsubsection}{\S#2#1#3}
\newcommand{\tick}{\includegraphics[height=1.5ex]{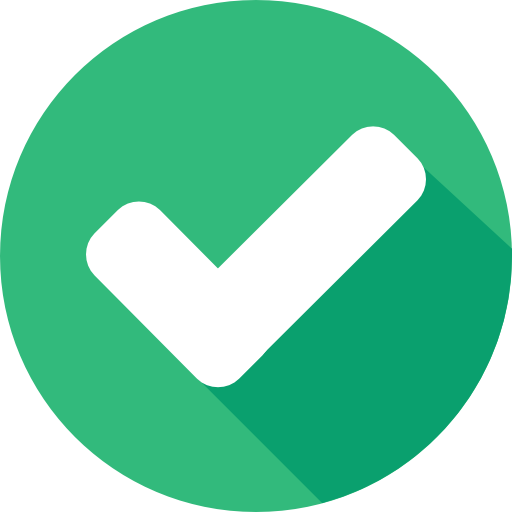}}
\newcommand{\cross}{\includegraphics[height=1.5ex]{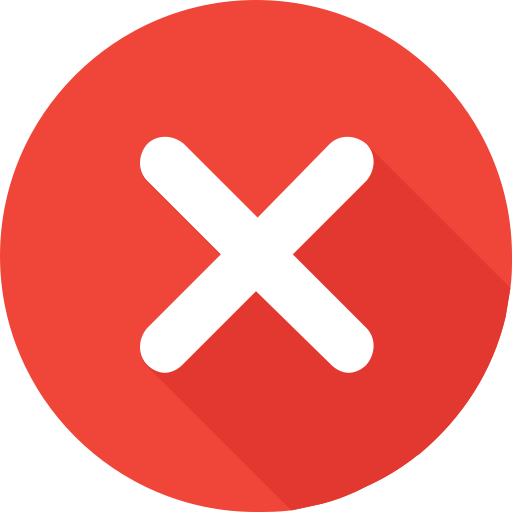}}

\newcommand{\name}[1]{\textsc{AIrchitect v2}\xspace}

\newcommand{\uov}[1]{{\textbf{\texttt{UOV}}}\xspace}

\begin{document}

\title{\name{}: Learning the Hardware Accelerator Design Space through Unified Representations  
}

\input{include/authors}

\maketitle

\begin{abstract}
Design space exploration (DSE) plays a crucial role in enabling custom hardware architectures, particularly for emerging applications like AI, where optimized and specialized designs are essential. With the growing complexity of deep neural networks (DNNs) and the introduction of advanced large language models (LLMs), the design space for DNN accelerators is expanding at an exponential rate. Additionally, this space is highly non-uniform and non-convex, making it increasingly difficult to navigate and optimize. Traditional DSE techniques rely on search-based methods, which involve iterative sampling of the design space to find the optimal solution. However, this process is both time-consuming and often fails to converge to the global optima for such design spaces. Recently, {\sc{AIrchitect} v1}, the first attempt to address the limitations of search-based techniques, transformed DSE into a constant-time classification problem using recommendation networks. However, {\sc{AIrchitect} v1} lacked generalizability and had poor performance in complex design spaces.  In this work, we propose {\sc{AIrchitect} v2}, a more accurate and generalizable learning-based DSE technique applicable to large-scale design spaces that overcomes the shortcomings of earlier approaches. Specifically, we devise an encoder-decoder transformer model that (a) encodes the complex design space into a uniform intermediate representation using contrastive learning and (b) leverages a novel unified representation blending the advantages of classification and regression to effectively explore the large DSE space without sacrificing accuracy. Experimental results evaluated on $10^5$ real DNN workloads demonstrate that, on average, {\sc{AIrchitect} v2} outperforms existing techniques by 15\% in identifying optimal design points. Furthermore, to demonstrate the generalizability of our method, we evaluate performance on unseen model workloads and attain a $1.7\times$ improvement in inference latency on the identified hardware architecture. Code and dataset are available at: \url{https://github.com/maestro-project/AIrchitect-v2}.
\end{abstract}


\input{contents/01-Introduction}

\input{contents/02-Background}

\input{contents/03-Methodology}

\input{contents/04-Evaluation}

\input{contents/06-Related-Works}
\input{contents/07-Conclusion}

\section*{Acknowledgment}
We thank Ananda Samajdar for their valuable assistance in helping set up {\sc AIrchitect v1} experiments for comparison and their insightful discussions on learning-based DSE. This work was funded in part by JUMP 2.0, a Semiconductor Research Corporation (SRC) program sponsored by DARPA.

\clearpage
\bibliographystyle{IEEEtran}
\bibliography{ref}

\end{document}

%% file: include/packages.tex
\usepackage{cite}
\usepackage{amsmath,amssymb,amsfonts}
\usepackage{graphicx}
\usepackage{textcomp}
\usepackage{xcolor}
\usepackage[a4paper, total={184mm,239mm}]{geometry}
\def\BibTeX{{\rm B\kern-.05em{\sc i\kern-.025em b}\kern-.08em
    T\kern-.1667em\lower.7ex\hbox{E}\kern-.125emX}}

\usepackage[]{hyperref}
\usepackage{xspace}
\usepackage[linesnumbered]{algorithm2e}
\usepackage{diagbox}
\RestyleAlgo{ruled}
\usepackage{graphicx}
\usepackage{textcomp}
\usepackage{xcolor}
\usepackage{caption}
\usepackage{threeparttable}
\usepackage{tabularx}
\usepackage{amsmath}
\usepackage{makecell}
\usepackage{multirow}
\usepackage{multicol}
\usepackage{pifont}
\usepackage{subfigure}
\usepackage{comment}

\usepackage{algpseudocode}

%% file: include/authors.tex
\author{

\IEEEauthorblockN{ Jamin Seo$^{*1}$, Akshat Ramachandran$^{*1}$,Yu-Chuan Chuang$^2$, Anirudh Itagi$^1$, Tushar Krishna$^1$}
\IEEEauthorblockA{$^1${Georgia Institute of Technology, Atlanta, USA}\\
$^2${National Taiwan University, Taipei, Taiwan}\\
$^1$\{akshat.r, jseo89, anirudh.itagi\}@gatech.edu, tushar@ece.gatech.edu}
$^2$frankchuang@access.ee.ntu.edu.tw
\thanks{$^*$Equal contribution.}



}

%% file: contents/01-Introduction.tex
\section{Introduction}
In the rapidly evolving domain of deep neural networks (DNNs), hardware acceleration \cite{ramachandran2024algorithm, kwon2018maeri, ramachandran2024microscopiq} is essential for the efficient deployment of models across a wide range of platforms, from cloud infrastructures to mobile and edge devices. However, the performance of DNN inference on hardware is dictated by the complex interaction between mapping strategies \cite{dosa} and allocated hardware resources \cite{airchitect}. This complexity leads to a vast and intricate design landscape, making it difficult to explore and optimize for peak performance. 

In the past, human experts have manually crafted the design choices based on their insights \cite{nvdla, chen2016eyeriss_jssc, du2015shidiannao}. Such endeavors not only require substantial time and resources but also may achieve sub-optimal solutions due to heuristics. Recently, motivated by the success of machine learning (ML) algorithms to perform complex tasks \cite{deng2009imagenet, xiao2018unified}, efforts are being made to automate DSE using ML techniques \cite{vaesa, airchitect, dosa}.

\begin{figure}[t]
    \centering
    \includegraphics[width=0.85\columnwidth]{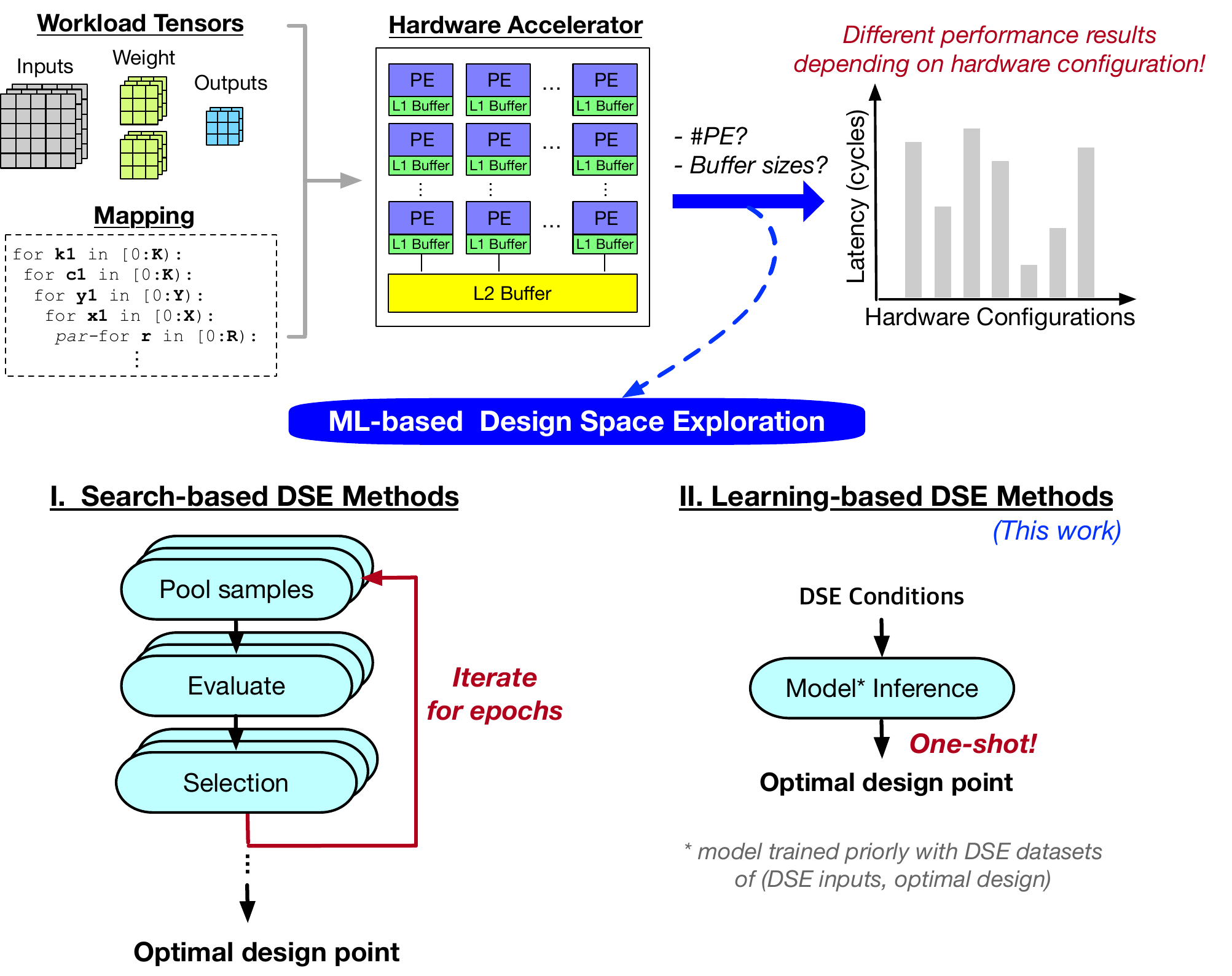}
    \caption{Different design choices yield a wide range of performance, necessitating automated design space exploration. (i) Search-based methods involve iterative exploration, (ii) learning-based methods enable one-shot inference. }
    \vspace{-15pt}
    \label{fig:dse_overview}
    
\end{figure}


Mainstream methodologies using ML in DSE automation commonly involve iterative searches of samples and often rely on black-box optimization techniques. For instance, ConfuciuX~\cite{confuciux} utilizes reinforcement learning (RL), Gamma~\cite{gamma}, DiGamma~\cite{digamma} apply genetic algorithm (GA), and Hasco~\cite{Xiao2021HASCOTA} employs Bayesian optimization (BO) for searching optimal hardware and/or mapping configurations. However, these techniques are often prohibitively time-consuming for large design spaces and the quality of identified designs is largely dependent on sampling efficiency \cite{gamma, vaesa}. 

To mitigate the inefficiencies of search-based DSE, recent techniques \cite{airchitect, feng2023gandse} propose a constant time optimization method by employing different DNN models trained/fine-tuned for DSE to predict the optimal hardware configuration. {\sc AIrchitect v1}~\cite{airchitect} proposed and demonstrated this idea on several hardware and mapping DSE tasks on systolic arrays~\cite{scalesim}, by training a multi-layer perceptron (MLP)-based classification model. Despite achieving good results on the systolic array design space, the accuracy and generality of this scheme are still less than expected (\cref{sec:results}). First, it did not address the non-uniform and long-tailed distribution of DSE data (\cref{sec:background_motivation}), which significantly impacts the learnability of DNN models. Second, modeling DSE as a classification-only problem significantly prunes the design space with a fixed number of labels, i.e. configuration, and restricts its scalability for larger DSE due to a commensurate increase in model size. And lastly, the design space explored by this technique is small and relatively simple.

  
\textbf{Contributions. }To address the above issues, we propose {\sc AIrchitect v2}, an enhanced version to enable more accurate, generalizable, and scalable learning-based DSE. We leverage contrastive learning to learn and encode the input feature representations into a uniform and smooth feature embedding space\footnote{In this work, we use intermediate representation and embedding space interchangeably.}. We also propose a novel unified representation namely, \textit{Unified Ordinal Vectors} \cite{ramachandran2023ntrans}, that enables joint classification and regression to leverage the unified benefits of both these techniques for DSE to overcome the limitation of the classification-only approach. Finally, we study the applicability of the proposed technique to a sufficiently complicated design space, the hardware resource allocation (for a given workload and mapping) on an accelerator modeled by MAESTRO~\cite{maestro}. 

Our key contributions can be summarized as follows:

\begin{itemize}[align=right,itemindent=2em,labelsep=2pt,labelwidth=1em,leftmargin=0pt,nosep]
    
    
    \item We propose \name{}, featuring (i) a novel encoder-decoder, multi-head self-attention-based model, (ii) contrastive learning approach for uniform and smooth embedding space, and (iii) a unified ordinal vector representation of output, combining classification and regression. (\cref{sec:method})

    
    \item Through extensive ablations and experiments, we demonstrate that AIrchitect v2 surpasses existing techniques by an average of $15\%$ in discovering optimal design points and achieves a $\sim 1.7\times$ improvement in inference performance on the predicted hardware architecture. (\cref{sec:results})

    \item We release our MAESTRO~\cite{maestro}-based DSE training dataset comprising of $10^5$ real DNN workloads along with our trained models to advance learning-based DSE research. 
\end{itemize}

%% file: contents/02-Background.tex
\section{Background and Motivation}
\label{sec:background_motivation}

\subsection{Learning-based DSE}
Training a DNN model that predicts the optimal design choice enables one-shot inference and mitigates sensitivity to sampling efficiency unlike search-based techniques (\autoref{fig:dse_overview}). {\sc AIrchitect v1}~\cite{airchitect} formulated several DSE tasks for systolic arrays as a classification problem, where its MLP model outputs probability distributions over labels, i.e. encoded design choices. The model is trained on a dataset of DSE input parameters and their corresponding optimal design choice, generated by the Scale-Sim~\cite{samajdar2020systematic} simulator. Given its effectiveness and advantages in the DSE domain, we expand this direction.

\subsection{Contrastive Learning}
Contrastive learning, widely adopted in self-supervised settings \cite{fu2022contrastive, cao2022synergistic} has proven to be effective in mitigating overfitting and improving generalization by regularization against negative samples. Recent work \cite{frumkin2023jumping, ramachandran2024clamp} has also demonstrated its benefits in smoothening the loss landscape. The core idea of contrastive learning, which is based on the infoNCE loss \cite{wang2023positive} is, to balance the learning of a data sample (anchor) by aligning the corresponding positive sample pairs and repulsing the negative sample pairs. As we shall demonstrate in \cref{sec:method}, by selecting positive samples from DSE data points belonging to the same class and aligning them in the embedding space, while simultaneously pushing away samples from different classes, contrastive learning promotes the creation of a more uniform embedding space and simultaneously combats the effects of long-tailed distributions \cite{li2022targeted}.

\begin{figure}[t!]
    \centering
    \includegraphics[width=0.5\textwidth]{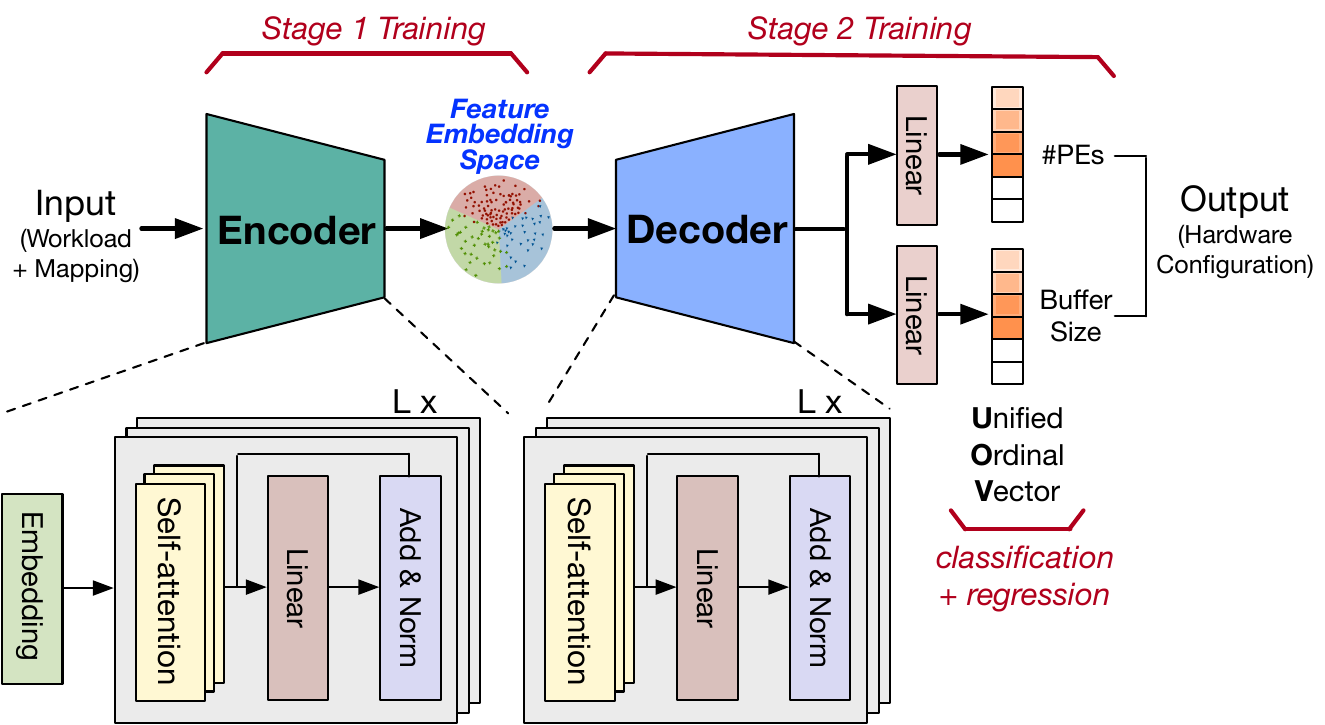}
    \caption{Overview of \name{}, highlighting (1) multi-head self-attention-based encoder and decoder structure, (2) latent embedding space improved by contrastive learning (3) \uov{} output representation combining classification and regression.}
    \label{fig:overview}
    \vspace{-5mm}
\end{figure}

\subsection{Motivation 1: Non-Uniform Performance Landscape}
As we can observe from \autoref{fig:challenge} (a), the normalized performance (latency) landscape, drawn from the DSE dataset introduced in \cref{sec:prob-formulation}, the distribution is highly non-uniform and non-convex. This makes it particularly challenging for search-based techniques to reach the global optimum due to a high probability of getting trapped within the multiple local minima \cite{frumkin2023jumping, ramachandran2024clamp}. Additionally, learning-based techniques also struggle to achieve good performance in the presence of such a non-uniform design space (\cref{sec:results}). For instance, even insignificant variations in the input features may cause the predictions to have large discrepancies since the model might not have converged to the global optimum. Furthermore, they are also prone to overfitting the training data because such a landscape can cause the model to fit too closely to the specific training data points, reducing its ability to generalize. In this work, we leverage contrastive learning to smoothen the landscape and encode it into a simpler and uniform embedding space.

\subsection{Motivation 2: Long-tailed Data Distribution}
\begin{figure}[t]
    \centering
        \includegraphics[width=0.9\columnwidth]{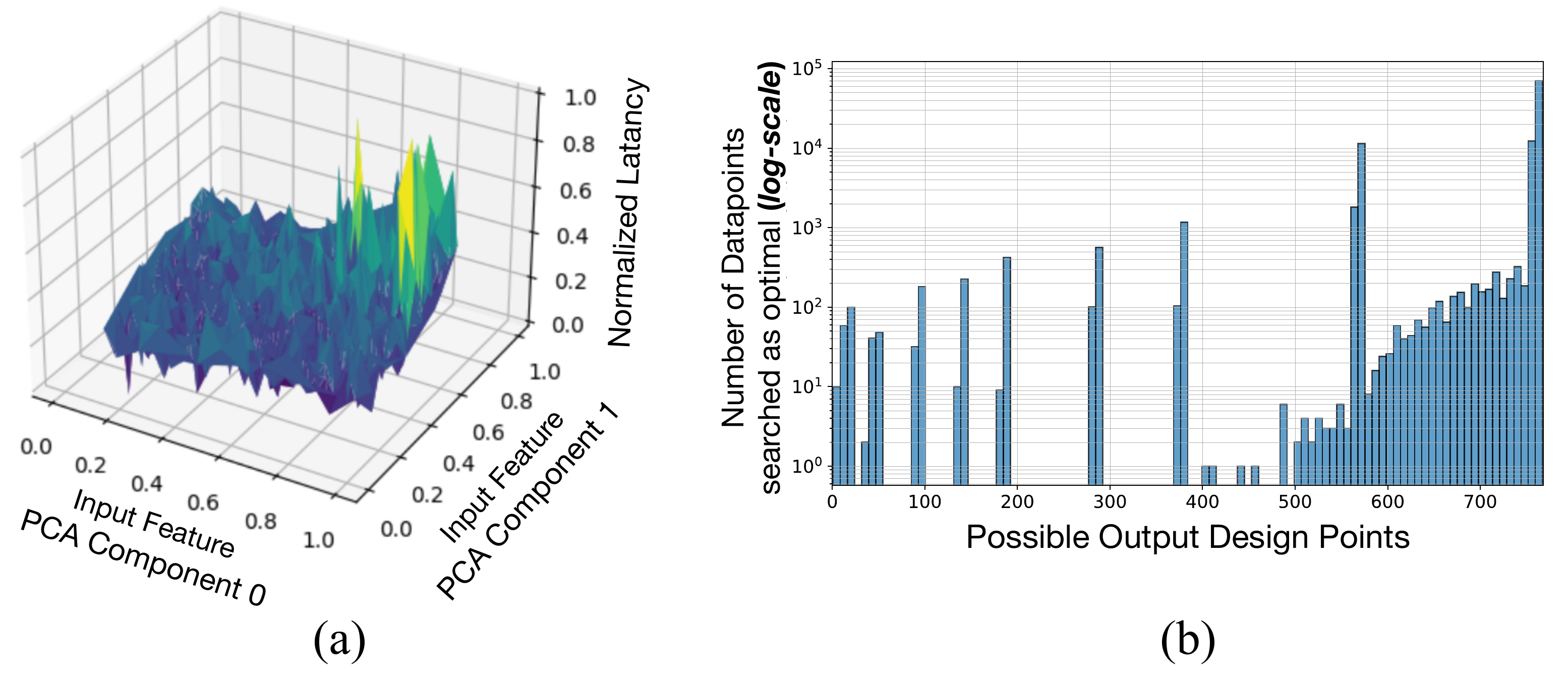}
    \caption{Prominent challenges on DSE dataset: (a) non-uniform and non-convex landscape (b) long-tailed distribution of data samples over labels. \textit{Drawn from the problem space in \ref{sec:prob-formulation} }}
    \label{fig:challenge}
    \vspace{-4mm}
\end{figure}

 \autoref{fig:challenge} (b), plots the number of data samples for each DSE output design point, based on the same dataset (\cref{sec:prob-formulation}) and using a strategy akin to that described in \cite{airchitect}. We observe that the DSE dataset is imbalanced and exhibits a long-tailed distribution \cite{li2022targeted}, where a small subset of output design points are favored by the majority of data samples, while many other design points are sparsely chosen. Such a data distribution poses a significant challenge for learning-based DSE techniques and impacts performance and generalizability (\cref{sec:results}). In this work, we extend the advantages of contrastive learning to address the long-tailed data distribution. By employing contrastive learning, the learnt embedding space converges to one where output design points are more uniformly distributed, reducing the imbalance in representation (see \autoref{fig:embedding}).

\subsection{Motivation 3: Classifications v/s Regression}
\label{sec:class_reg}
Depending on how the optimal DSE output is determined, we categorize existing DSE techniques into two broad categories: \textit{classification} and \textit{regression}. Classification-based techniques, such as {\sc AIrchitect v1} \cite{airchitect}, partition the design space into a set of predefined classes or labels, each representing a distinct design configuration. It predicts the optimal design point by selecting the most appropriate class from this fixed set, simplifying the search process and better constraining the search space. However, this approach can limit flexibility and scalability when dealing with large or complex design spaces. 

We take the liberty of categorizing techniques as regression-based when they are capable of predicting DSE output hardware configurations as continuous values, rather than selecting from a predefined set of discrete options as described above. Consequently, under this definition, all search-based techniques \cite{confuciux, gamma, digamma, Xiao2021HASCOTA, dosa, vaesa} can be interpreted as regression-based. Regression-based techniques \cite{feng2023gandse, dosa, vaesa} are highly scalable since increasing design space size or complexity does not necessitate an increase in model size \cite{feng2023gandse, vaesa}. However, with large and complex design spaces, these methods result in an unconstrained learning problem \cite{roelofs2019meta}, which can greatly impact accuracy and increase the risk of overfitting. In this work, we propose a novel representation \uov{}, or \textit{Unified Ordinal Vectors} that can leverage the unified benefits of both these techniques while mitigating their specific drawbacks.

%% file: contents/03-Methodology.tex
\section{\name{}}\label{sec:method}


\subsection{DSE Problem Formulation}\label{sec:prob-formulation} 

To delve into the aforementioned challenges in DSE, we select the following task as our target scenario for exploration: hardware resource assignment on MAESTRO\cite{maestro}-modeled accelerator. This problem, previously also explored by ConfuciuX~\cite{confuciux} using RL-based search, has been shown to be a sufficiently complex design space.

We translate the DSE task into a learnable formulation, by encoding the design parameters following \autoref{tab:problem-def}, modified from \cite{confuciux}. As the DSE inputs, the tenser dimensions for GEMM operation ($M,N,K$) are numerical integer values, while dataflow is categorical data chosen among: weight stationary \cite{nvdla}, output stationary \cite{du2015shidiannao}, and row stationary\cite{chen2016eyeriss_jssc}. 

The output is the optimal hardware resources for the given per-layer inputs, configured as the number of PEs and L2 buffer size, while L1 buffer size is fixed following the search assumptions in~\cite{confuciux}.
The dataset is generated by executing ConfuciuX~\cite{confuciux} on the randomized input parameters, with the optimization metric (i.e. reward) set as latency. 

\begin{figure}[t]
    \centering
    \includegraphics[width=0.6\columnwidth]{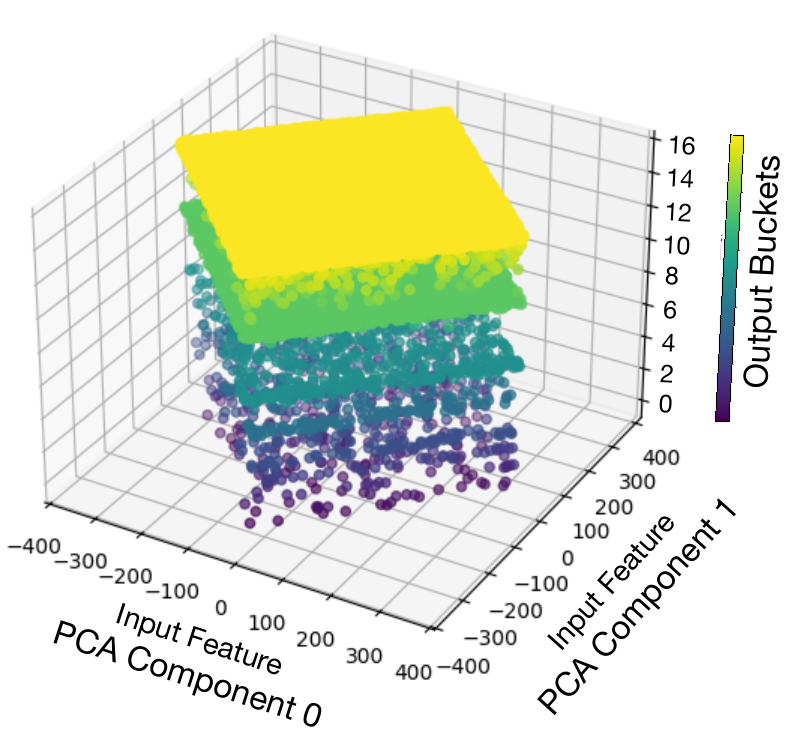}
    \caption{Complexity of the problem space from \autoref{tab:problem-def}, visualizing the input features ($xy$-plane, processed with PCA) and output configuration ($z$-axis,  plotted into \uov{} buckets). This justifies the need for sophisticated model architecture. }
    \label{fig:prob-space}
    \vspace{-4mm}
\end{figure}

This DSE task forms a large design space of complexity, $O(10\textsuperscript{9})$ 
derived as the product of input feature dimensions from \autoref{tab:problem-def}.
\autoref{fig:prob-space} visualizes its significant complexity, exhibiting irregular and non-trivial characteristics that hardly suffice with simple techniques such as decision trees or support vector machines \cite{airchitect}. Moreover, the dataset shows a highly non-uniform performance landscape as well as imbalanced data sample distribution as highlighted in ~\autoref{fig:challenge}.

\input{tables/problem-def}

\subsection{{\sc AIrchitect v2} Overview}
\label{sec:overview}
We present an overview of {\sc AIrchitect v2} framework in \autoref{fig:overview}.  To effectively learn the complex DSE space, we design an encoder-decoder transformer-based model architecture following the structure in \cite{vaswani2017attention} (see \cref{sec:results} for reasoning). {\sc AIrchitect v2} takes as input workload specifications as outlined in \autoref{tab:problem-def} and outputs optimized hardware design configurations that are geared towards improving overall latency and/or energy. 

Both the encoder and decoder have identical and complementary structures, consisting of $L$ layers of stacked self-attention blocks, a feed-forward network, and a downsampling (encoder) / upsampling (decoder) units \cite{vaswani2017attention}. The decoder also has two \uov{} heads (explained later) corresponding to the two hardware design configurations predicted by the framework.

The encoder and decoder in the {\sc AIrchitect v2} framework decompose the hardware DSE learning and prediction process into two distinct stages. In stage 1 (\cref{sec:stage_1}), the encoder is responsible for learning to construct a uniform and smooth intermediate representation of the input design space, and during prediction, identifies a point in this learned embedding space that closely approximates the input specifications. In stage 2 (\cref{sec:stage_2}), the decoder learns to process the identified point in the intermediate representation and finally predict the optimal design configuration via \uov{}.



\subsection{\name{} Stage 1}
\label{sec:stage_1}

\begin{figure}[t]
    \centering
        \includegraphics[width=0.7\columnwidth]{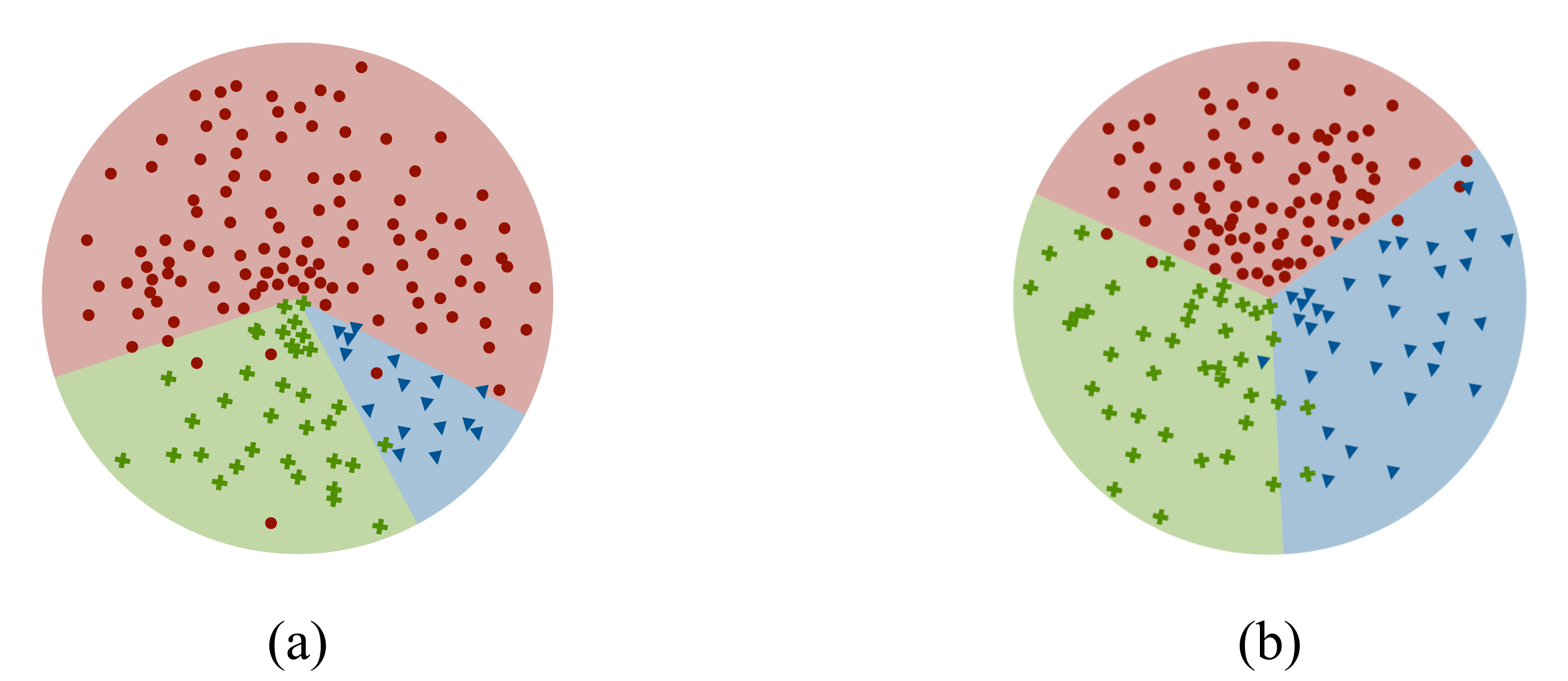}
        \vspace{-3mm}
    \caption{Visualization of embedding space (a) without contrastive learning and b) with contrastive learning. Employing contrastive learning results in a uniform embedding space. Different colors represents different classes of data samples.}
    \label{fig:embedding}
    \vspace{-5mm}
\end{figure}

The goal of stage 1, i.e. encoder, is to generate a uniform and smooth intermediate representation of the input design space. To guide the encoder in efficiently learning this intermediate space, we leverage a combined objective consisting of the contrastive term and performance prediction term.

\noindent\textbf{Contrastive Learning. }As pointed out in \cref{sec:background_motivation}, contrastive learning enables the encoder to learn to create a uniform and smooth embedding space by aligning positive samples together while simultaneously pushing away negative samples. In the context of stage 1 training, for each workload configuration within an input batch (anchor), positive and negative samples correspond to configurations that belong to the same and different \uov{} buckets, respectively. Inspired by \cite{ramachandran2024clamp, fu2022contrastive}, we adopt the infoNCE loss variant \cite{fu2022contrastive} of contrastive loss, and augment it to balance the positive and negative samples. The contrastive objective can be defined as,
\begin{equation}
\label{equation:contrastive}
    \mathcal{L}^C = -\log \frac{ \sum_{p+}\exp(\lambda^p \cdot \lambda^{p+} / \tau)}{ \sum_{p+} \exp (\lambda^p \cdot \lambda^{p+} / \tau) + \sum_{p-} \exp(\lambda^p \cdot \lambda^{p-} / \tau)}
\end{equation}   

where, $\lambda$ is the output embedding representation from the encoder, and p, p+, and p- are the anchor, positive, and negative samples, respectively. $\tau$ is empirically determined to be 0.4. 

\noindent \textbf{Performance Predictor. }Training the encoder with a vanilla contrastive objective will create an embedding space with no semantic meaning \cite{vaesa}. Therefore, we augment the training objective with an L1-based performance prediction loss, $\mathcal{L}^{perf}$ to add semantic information to the learnt embedding. The performance here is the optimization goal of DSE, e.g. latency. This design is motivated by earlier work \cite{vaesa, gomez2018automatic}, that emphasizes the influence of performance predictors in organizing the embedding space.  
    

The final stage-1 objective is, $\mathcal{L}^{stage1} = \mathcal{L}^C + \mathcal{L}^{perf}$. During stage 1 training, the encoder is trained with ${L}^{stage1}$ as the objective, enabling the encoder to learn the embedding space that keeps similar DSE samples close while distancing dissimilar ones (\autoref{fig:embedding}) and incorporate enriched semantic information that aligns with the resulting performance. This contributes to the formation of a uniform and smooth intermediate representation space that is also robust to the imbalanced and long-tailed data distribution.

\subsection{\name{} Stage 2}
\label{sec:stage_2}

Once stage 1 training is complete, we train the decoder in stage 2 while keeping the encoder's weights fixed to prevent the backpropagation of gradients. In this stage, the decoder learns to predict the optimal DSE hardware configuration given a point in the embedding space identified by the encoder. Unlike previous works \cite{airchitect, feng2023gandse, vaesa} the decoder is augmented to predict our proposed \uov{} through \uov{} heads (\autoref{fig:overview}) which are simple feed-forward layers that learn to predict \uov{}s guided by the stage 2 objective. 
Since the DSE space explored in this study has two configurations, i.e. the number of PEs and buffer size, the decoder has two \uov{} heads.

\noindent \textbf{Unified Ordinal Vectors }(\uov{}). The \uov{} representation \cite{ramachandran2023ntrans} scheme enables embedding the large-scale classification labels into reduced-size and scalable representation via "bucketization". Based on the scheme, the model jointly and implicitly predicts the classification \textit{bucket} while regressing to the actual DSE configuration within each bucket. The classification bucket consists of \textit{ranges} of DSE points. We employ Space Increasing Discretization \cite{miao2023occdepth} for the given DSE space and obtain $K$ discretized buckets, $\Lambda = \{r_0, r_1, ..., r_{K-1} \}$. The higher the value of K, the lower the range of DSE points covered by each bucket. Following \autoref{alg:enc}, any DSE configuration $D$ can be encoded as a K-length \uov{} such that,

\begin{itemize}
    \item Assuming $D$ lies in bucket $r_n$
    \item Bucket values preceding $r_n$ are non-zero and monotonically increasing (\autoref{alg:enc} \autoref{line:nz}) 
    \item Bucket values following $r_n$ are zero (\autoref{alg:enc} \autoref{line:z}) 
\end{itemize}

\begin{figure}[t]
    \vspace{-2mm}
    \centering
    \includegraphics[width=0.9\columnwidth]{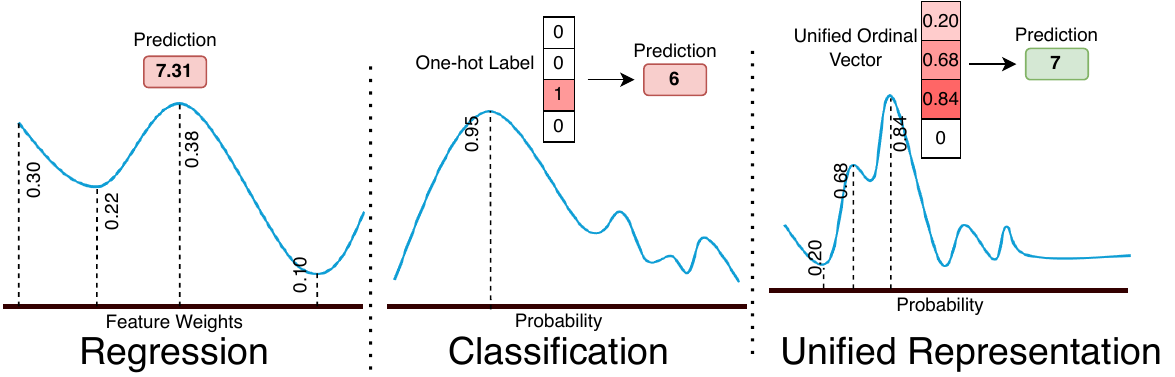}
    \vspace{-2mm}
    \caption{Visualization of how the same configuration is represented for regression, classification, and the proposed \uov{}.}
    \label{fig:uov}
    \vspace{-5mm}
\end{figure}

As a result, the final \uov{} ($O$) of a given $D$ is,

\begin{equation}
\{O_i\}_{i=0}^{K-1}=
\begin{cases}
  1-f(|D - r_i|), & \text{if } D \geq r_i \\
  0, & \text{otherwise}
\end{cases}
\end{equation}

where $f$ can be any choice of monotonically increasing function (we select the exponential function in \autoref{alg:enc}). \autoref{fig:uov} intuitively visualize this process, and decoding the \uov{} back to the actual DSE configuration is the exact reverse of \autoref{alg:enc}. Our proposed unified ordinal representation captures essential ordering information and is well-suited for the model to learn and predict thanks to its regular structure. For evaluation, we empirically set $K$ as 16 (see \cref{sec:ablations}).

\textbf{Decoder Training. }The decoder and \uov{} heads are trained with the same data used for stage 1. 
By leveraging the transformer decoder as the backbone, each \uov{} head is trained to predict the corresponding hardware configuration by unifying (1) classification to identify range buckets and (2) regression for finer-granularity search within the bucket.

\textbf{Unification Loss. }We make use of the \textit{Unification Loss} ($L_o$) as the primary training loss. Due to the nature of the \uov{}, we adopt a loss function similar to \cite{li2020generalized} to guide stage 2 training,  

\vspace{-1mm}
\begin{equation}
    \hspace{-2pt}L_o = \sum_{i=0}^{{K-1}}\hspace{-2pt}
    \begin{cases}
        \alpha|q_i-u_i|^\gamma BCE(u_i,q_i),& \hspace{-5pt} \text{if } q_i > 0 \\ 
        (1-\alpha)u_i^\gamma BCE(u_i,q_i),& \hspace{-5pt} \text{otherwise}
    \end{cases}
\end{equation}
\vspace{-2mm}
\begin{equation}
 BCE(u_i,q_i)= -q_ilog(u_i)-(1-q_i)log(1-u_i)
\end{equation}
where, $BCE$ corresponds to Binary Cross-Entropy, $u,q$ are the predicted and ground-truth \uov{}s 
respectively, and $\alpha=0.75$, $\gamma=1$ are empirically determined. This formulation for the unification loss penalizes the predictions buckets farther from the ground-truth bucket more heavily than those closer to the ground-truth. In addition, it penalizes the actual likelihood/regression within a predicted bucket to simultaneously ensure accurate bucket prediction and correct estimation of the actual DSE point within the bucket.

\begin{algorithm}[t]
\caption{Ordinal Encoding}\label{alg:enc}
\SetKwInOut{KwIn}{Input}
\SetKwInOut{KwOut}{Output}
\SetKwComment{Comment}{/* }{ */}
\KwIn{Ground-truth DSE-config $ D_{gt} \in \mathbb{R}$, $K$ buckets produced $\Lambda = \{r_0, r_2, ... , r_{K-1}\}$, $r_i\in \mathbb{R}$}

\KwOut{Ordinal vectors $\{O_i \in \mathbb{R}\}_{i=1}^K$}

    \For{$i = 0$ to $K-1$}
    {
        \If{$D \geq r_i$}
        { 
            $O_i \gets 1-\exp^{-|{D - r_i}|}$\;  \label{line:nz}
        }

        \Else
        {
            $O_i \gets 0 $\; \label{line:z}
        } 
    }
\Return $\{ O_i \}_{i=1}^K$
\end{algorithm}

\subsection{\name{} Deployment Pipeline}\label{sec:deploy-method}
\name{} is trained and inferred on a per-layer basis, recommending the optimal hardware resources for single-layer execution. For model-level deployment, we mention two methods (which apply to any 
 general layer-granularity DSE). 

Given a model with $N$ layers, $\textbf{M}=\{L_0, L_1, ..., L_N\}$, assume \name{} has recommended $\textbf{HW}=\{HW_{L_0}, HW_{L_1}, ..., HW_{L_N}\}$ for each layer. We can determine the final hardware configuration $HW_{M}$ from either:

\textit{\underline{Method 1.}} For each $HW_{L_i}$ in $\textbf{HW}$, estimate the model-wise latency (we use MAESRO~\cite{maestro} in this work) across all layers in \textbf{$M$}. Select the $HW_{L_i}$ that yields the minimum as $HW_{M}$. 

\textit{\underline{Method 2.}} Identify the bottleneck layer $L_n$ among $L_i$ that results in the largest latency when executed on its recommended $HW_{L_i}$. Choose the $HW_n$ as $HW_{M}$.

We demonstrate \textit{\underline{Method 1}} on representative models in ~\cref{sec:model-eval}, with per-layer DSE from \name{} and baselines, to highlight the practical effectiveness of our approach.

%% file: tables/problem-def.tex
\begin{table}[t]
\scriptsize
\centering
\setlength{\abovecaptionskip}{0pt}
\renewcommand{\arraystretch}{1.5}
\caption{Input and output formulations for MAESTRO~\cite{maestro}-based hardware resource allocation, modified from \cite{confuciux}}
\label{tab:problem-def}
  \begin{threeparttable}[b]
\begin{tabular} {|c|p{6cm}|}
\hline
\multicolumn{1}{|c|}{\textbf{}} &
\multicolumn{1}{c|}{\textbf{Features (size)}} 
\\
\hline

\hline
Input & M (256), N (1677), K (1185), dataflow (3)  \\
\hline
Output & PE (64), buffer size (12)  \\

\hline
\end{tabular}
 \begin{tablenotes}
     \item[] - Assuming GEMM operation $(M,K) \times (K,N) = (M,N$) 
 \end{tablenotes}
 \begin{tablenotes}
    \item[] - dataflow=choice among [\cite{nvdla}, \cite{du2015shidiannao}, \cite{chen2016eyeriss_jssc}] 
 \end{tablenotes}
 
 \end{threeparttable}
\end{table}


%% file: contents/04-Evaluation.tex
\section{Results and Discussion}
\label{sec:results}

\subsection{Experimental Setup}\label{sec:exp-setup}

\noindent \textbf{Implementation. } The \name{} framework is implemented in Pytorch and evaluated on the DSE task introduced in \cref{sec:prob-formulation} with a dataset consisting of 100K samples, split into 80K for training and 20K for testing. We train \name{}'s stage 1 and stage 2 individually for 500 and 100 epochs, respectively. All experiments were conducted on a single NVIDIA H100 GPU. The access to the dataset, scripts for training, and the pre-trained encoder and decoder models are provided in \url{https://github.com/maestro-project/AIrchitect-v2}.

\noindent \textbf{Baselines. } We compare \name{} framework against existing SoTA learning-based techniques, including the MLP-based {\sc AIrchitect v1}~\cite{airchitect}, generative adversarial network GANDSE~\cite{feng2023gandse} and variational autoencoder VAESA \cite{vaesa} combined with a search-based technique (BO), which are trained and evaluated on the same dataset (\cref{sec:method}) for fair evaluation. 

\vspace{-1mm}

\input{tables/results_comparison}
\subsection{Layer-level Prediction Accuracy}
As shown in \autoref{table:accuracy}, \name{} demonstrates considerable improvement in prediction accuracy over other baselines. In particular, the shallow MLP model and classification head used in {\sc AIrchitect v1} cause significant overfitting to the training data and inability to handle the complexities of the DSE landscape and data distribution, resulting in the lowest accuracy of 77.60\%. Although GANDSE achieves higher accuracy than {\sc AIrchitect v1}, it is still limited by the large unconstrained learning problem due to its generative approach, impacting the quality of DSE outputs. In contrast, \name{} achieved a notably high accuracy of 91.17\%, benefiting from our solutions outlined in ~\cref{sec:overview}, ~\cref{sec:stage_1}, and ~\cref{sec:stage_2}. 

\vspace{-1mm}
\subsection{Model-level Deployment Evaluation}\label{sec:model-eval}

\begin{figure}[t]
    \centering
    \includegraphics[width=\columnwidth]{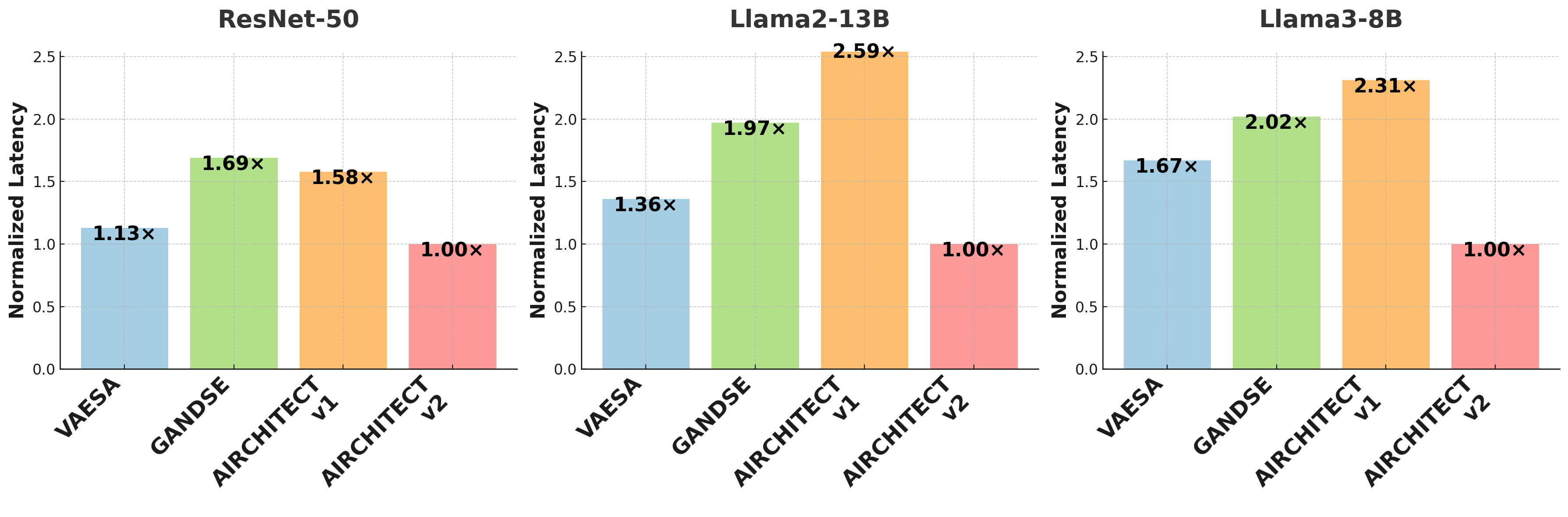}
    \vspace{-4mm}
    \caption{DSE performance (latency) comparison of different techniques on popular DNNs and LLMs. Normalized to \name{} (red). Latency is estimated using MAESTRO~\cite{maestro}. }
    \label{fig:model_performance}
    \vspace{-4.5mm}
\end{figure}

We further assess the performance in practical model deployment on representative DNN and LLM models~\cite{resnet50, touvron2023llama, dubey2024llama}, \textit{which were never seen during training}. \autoref{fig:model_performance} compares the model-level latency achieved by various DSE techniques mentioned in ~\cref{sec:exp-setup}. \name{} consistently outperforms others across workloads in identifying the hardware configuration with the lowest latency. Particularly, {\sc AIrchitect v1} and GANDSE \cite{feng2023gandse} achieve poor performance due to overfitting and solutions being trapped in local optima, as they lack addressing the non-uniform and non-convex DSE performance landscape. VAESA with BO \cite{vaesa} is the only method that achieves performance close to ours, as it is able to construct a continuous and low-dimensional latent space through a variational autoencoder (VAE). However, as demonstrated in \cref{sec:ablations}, the embedding space generated through contrastive learning is superior to VAE.

\vspace{-1mm}
\subsection{Ablations}\label{sec:ablations}
\vspace{-1mm}

\noindent \textbf{Encoder Training Loss.} We investigate the impact of contrastive loss and performance prediction loss to the training, as shown in \autoref{table:stage1_ablation}. Without both objectives (and using only an L2-loss term), the model struggles to handle the non-uniform DSE landscapes and skewed data distributions, resulting in low accuracy similar to {\sc AIrchitect v1}. Incorporating contrastive loss significantly alleviates these issues, leading to a substantial increase in prediction accuracy (+10.54\%). The addition of performance prediction loss further enhances learning (+1.2\%) by providing semantic information to the learnt feature embedding, achieving the highest final prediction accuracy.
\begin{figure}[t]
    \centering
    \includegraphics[width=0.9\columnwidth]{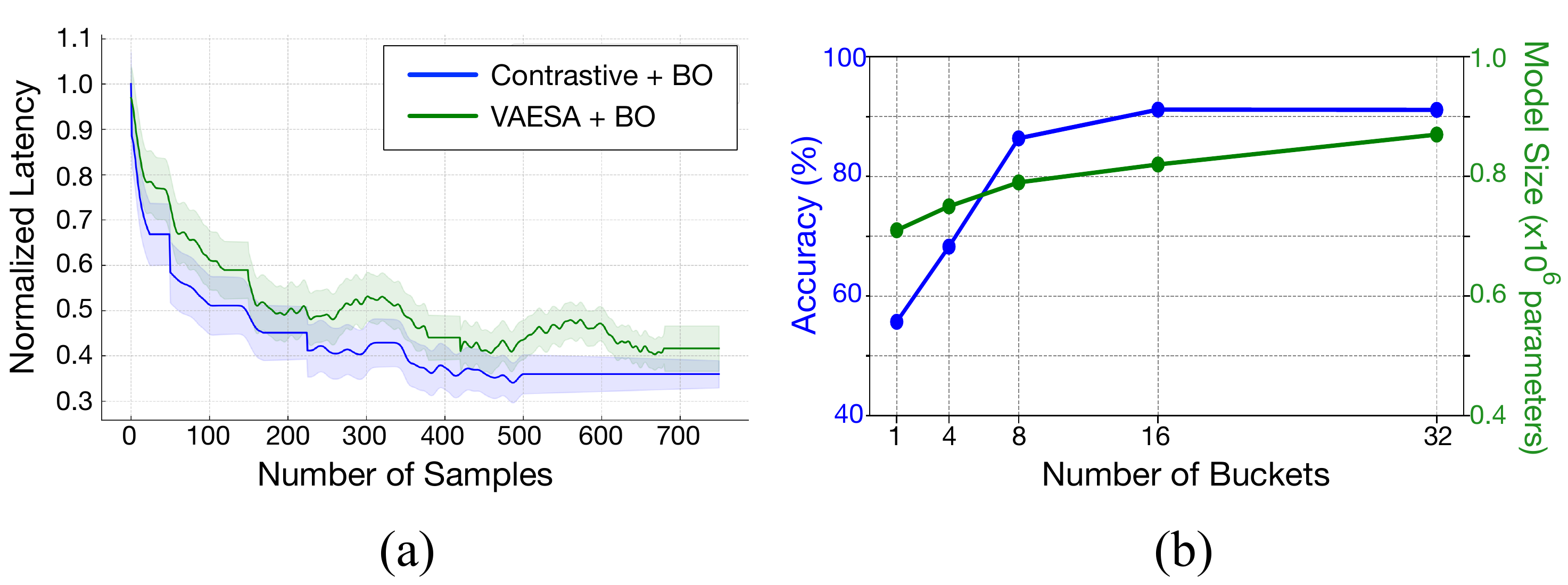}
    \vspace{-2mm}
    \caption{(a) DSE performance comparison of searching using BO on the contrastive embedding v/s VAE-generated embedding \cite{vaesa} for a Llama2-7B model. (b) Impact of number of \uov{} intervals within each bucket, on accuracy and model size.}
    \label{fig:cl-uov-ablation}
    \vspace{-5.5mm}
\end{figure}

\noindent \textbf{Impact of Contrastive learning. }
To study the effectiveness of the proposed contrastive learning on DSE feature embeddings, we further evaluate the constructed embedding space. Following \cite{vaesa}, we adopt BO to search from both the embedding space constructed by contrastive learning and the VAE-generated latent space \cite{vaesa}. We train a separate decoder for each technique that converts a point from the latent space into a hardware configuration, and estimate the corresponding latency using MAESTRO \cite{maestro}. As observed in \autoref{fig:cl-uov-ablation} (a), searching within our contrastive embedding space leads to significantly faster convergence and lower latency compared to the VAE-generated embedding space, implying a more uniform and smoother performance landscape.
\begin{figure}[t]
    \centering
    \includegraphics[width=0.9\columnwidth]{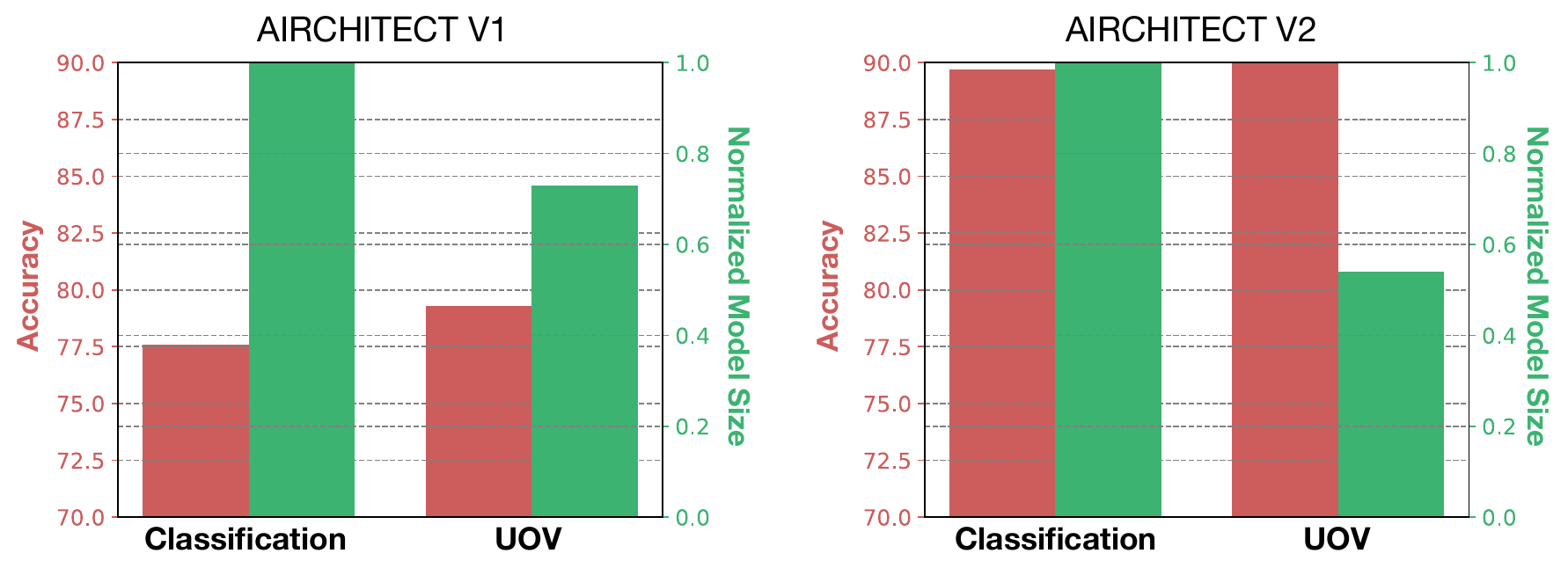}
    \vspace{-2mm}
    \caption{ Effect of \uov{} on prediction accuracy(red) and model size(green), for {\sc AIrchitect v1} and \name{}.
    }
    \label{fig:uov_cls}
    \vspace{-5.5mm}
\end{figure}

\noindent \textbf{\uov{} v/s Classification. } \autoref{fig:uov_cls} compares the effectiveness of the proposed \uov{} formulation against the conventional classification approach, for both {\sc AIrchitect v1} and \name{}. We observe that in both scenarios, \uov{} formulation improves prediction performance because classification overly discretizes and constrains the design space, while \uov{} combines the benefits of classification and regression, enabling fine-grained prediction. Additionally, \uov{} significantly reduces model size, which highlights its applicability in larger design spaces. By demonstrating the advantages of \uov{} for two different techniques, we show that it is not specific to \name{} and can be adapted for similar benefits in other methods.

\noindent \textbf{\uov{} Hyperparameter Evaluation. } Increasing the number of \uov{} buckets improves accuracy through finer-granularity prediction or reduced discretization, but also increases model size due to the larger output vector. As shown in \autoref{fig:cl-uov-ablation} (b), model size (green) grows along with the number of buckets, while accuracy (blue) begins to saturate beyond 16 buckets. We select 16 \uov{} buckets for our DSE learning to achieve the optimal trade-off between accuracy and model size. Notably, as the number of buckets increases, the problem shifts toward classification, while a single bucket reverts it to regression!

%% file: tables/results_comparison.tex
\begin{table}[!t]
\begin{minipage}{0.45\columnwidth}
		\scriptsize
		\caption{{\sc AIrchitect v2} stage 1 ablations.}
		\label{table:stage1_ablation}
		\centering
		\begin{tabular}{ccc}
\Xhline{2\arrayrulewidth}
$\mathcal{L}^C$ & $\mathcal{L}^{perf}$ & Accuracy (\%) \\
\Xhline{1\arrayrulewidth}
\cross & \cross & 79.43 \\
\cross & \tick & 81.27 \\
\tick & \cross & 89.97 \\
\tick & \tick & \textbf{91.17} \\
 \Xhline{2\arrayrulewidth}
\end{tabular}
	\end{minipage} \hfill
	\begin{minipage}{0.55\columnwidth}
	\scriptsize\addtolength{\tabcolsep}{-5pt}
		\caption{Comparison with other learning-based techniques.}
		\label{table:accuracy}
		\centering
		\begin{tabular}{cc}
\Xhline{2\arrayrulewidth}
Method & Accuracy (\%) \\
\Xhline{2\arrayrulewidth}
GANDSE \cite{feng2023gandse} & 84.39 \\
{\sc AIrchitect v1} \cite{airchitect} & 77.60 \\
\textbf{{\sc AIrchitect v2} (Ours)} & \textbf{91.17} \\
 \Xhline{2\arrayrulewidth}
\end{tabular}
\end{minipage}
\vspace{-4mm}
\end{table}

%% file: contents/06-Related-Works.tex
\section{Related Works}
\noindent \textbf{Search-based Optimizations Approaches.} \cite{gamma, digamma} utilize GA over genomes encoding design points. \cite{confuciux} uses RL for coarse-grained search, followed by GA for fine-tuning. \cite{Xiao2021HASCOTA} performs a two-step optimization combining multi-objective BO with Q-learning algorithms. 
\cite{mindmapping} employs a differentiable surrogate model to guide sampling via input gradients. 

\noindent \textbf{Supervised Learning-based Approaches.} 
\cite{airchitect} trains an MLP model to predict optimal design choices on systolic arrays, framing the DSEs as a classification problem. \cite{feng2023gandse} trains a GAN that generates design points to meet user-specified objectives, in a higher-dimensional design space. \cite{vaesa} focuses on DSE feature embedding space by training a variational autoencoder.


%% file: contents/07-Conclusion.tex
\section{Conclusion}
Current supervised learning-based DSE studies insufficiently addressed DSE-specific challenges in the dataset and learning formulation. In this paper, we propose \name{}, which employs an encoder-decoder transformer model to learn the complicated DSE space and leverages contrastive learning and \uov{} representation to tackle the non-uniform embedding space and long-tailed data distribution. \name{} improved prediction accuracy on the test set by $\sim$15$\%$ over competing baselines, while also achieving $1.7\times$ higher performance of identified designs on unseen workloads.

%% file: main.bbl
\begin{thebibliography}{10}
\providecommand{\url}[1]{#1}
\csname url@samestyle\endcsname
\providecommand{\newblock}{\relax}
\providecommand{\bibinfo}[2]{#2}
\providecommand{\BIBentrySTDinterwordspacing}{\spaceskip=0pt\relax}
\providecommand{\BIBentryALTinterwordstretchfactor}{4}
\providecommand{\BIBentryALTinterwordspacing}{\spaceskip=\fontdimen2\font plus
\BIBentryALTinterwordstretchfactor\fontdimen3\font minus \fontdimen4\font\relax}
\providecommand{\BIBforeignlanguage}[2]{{%
\expandafter\ifx\csname l@#1\endcsname\relax
\typeout{** WARNING: IEEEtran.bst: No hyphenation pattern has been}%
\typeout{** loaded for the language `#1'. Using the pattern for}%
\typeout{** the default language instead.}%
\else
\language=\csname l@#1\endcsname
\fi
#2}}
\providecommand{\BIBdecl}{\relax}
\BIBdecl

\bibitem{ramachandran2024algorithm}
A.~Ramachandran, Z.~Wan, G.~Jeong, J.~Gustafson, and T.~Krishna, ``Algorithm-hardware co-design of distribution-aware logarithmic-posit encodings for efficient dnn inference,'' \emph{arXiv preprint arXiv:2403.05465}, 2024.

\bibitem{kwon2018maeri}
H.~Kwon, A.~Samajdar, and T.~Krishna, ``Maeri: Enabling flexible dataflow mapping over dnn accelerators via reconfigurable interconnects,'' \emph{ACM SIGPLAN Notices}, vol.~53, no.~2, pp. 461--475, 2018.

\bibitem{ramachandran2024microscopiq}
A.~Ramachandran, S.~Kundu, and T.~Krishna, ``Microscopiq: Accelerating foundational models through outlier-aware microscaling quantization,'' \emph{arXiv preprint arXiv:2411.05282}, 2024.

\bibitem{dosa}
\BIBentryALTinterwordspacing
C.~Hong, Q.~Huang, G.~Dinh, M.~Subedar, and Y.~S. Shao, ``Dosa: Differentiable model-based one-loop search for dnn accelerators,'' in \emph{Proceedings of the 56th Annual IEEE/ACM International Symposium on Microarchitecture}, ser. MICRO '23.\hskip 1em plus 0.5em minus 0.4em\relax New York, NY, USA: Association for Computing Machinery, 2023, p. 209–224. [Online]. Available: \url{https://doi.org/10.1145/3613424.3623797}
\BIBentrySTDinterwordspacing

\bibitem{airchitect}
\BIBentryALTinterwordspacing
A.~Samajdar, J.~M. Joseph, M.~Denton, and T.~Krishna, ``Airchitect: Learning custom architecture design and mapping space,'' 2021. [Online]. Available: \url{https://arxiv.org/abs/2108.08295}
\BIBentrySTDinterwordspacing

\bibitem{nvdla}
NVIDIA, ``Nvdla deep learning accelerator,'' \url{http://nvdla.org}, 2017.

\bibitem{chen2016eyeriss_jssc}
Y.-H. Chen, T.~Krishna, J.~S. Emer, and V.~Sze, ``Eyeriss: An energy-efficient reconfigurable accelerator for deep convolutional neural networks,'' \emph{IEEE Journal of Solid-State Circuits}, vol.~52, no.~1, pp. 127--138, 2016.

\bibitem{du2015shidiannao}
Z.~Du, R.~Fasthuber, T.~Chen, P.~Ienne, L.~Li, T.~Luo, X.~Feng, Y.~Chen, and O.~Temam, ``Shidiannao: Shifting vision processing closer to the sensor,'' in \emph{International Symposium on Computer Architecture (ISCA)}, 2015.

\bibitem{deng2009imagenet}
J.~Deng, W.~Dong, R.~Socher, L.-J. Li, K.~Li, and L.~Fei-Fei, ``Imagenet: A large-scale hierarchical image database,'' in \emph{2009 IEEE conference on computer vision and pattern recognition}.\hskip 1em plus 0.5em minus 0.4em\relax Ieee, 2009, pp. 248--255.

\bibitem{xiao2018unified}
T.~Xiao, Y.~Liu, B.~Zhou, Y.~Jiang, and J.~Sun, ``Unified perceptual parsing for scene understanding,'' in \emph{Proceedings of the European conference on computer vision (ECCV)}, 2018, pp. 418--434.

\bibitem{vaesa}
Q.~Huang, C.~Hong, J.~Wawrzynek, M.~Subedar, and Y.~S. Shao, ``Learning a continuous and reconstructible latent space for hardware accelerator design,'' in \emph{2022 IEEE International Symposium on Performance Analysis of Systems and Software (ISPASS)}, 2022, pp. 277--287.

\bibitem{confuciux}
S.-C. Kao, G.~Jeong, and T.~Krishna, ``Confuciux: Autonomous hardware resource assignment for dnn accelerators using reinforcement learning,'' in \emph{2020 53rd Annual IEEE/ACM International Symposium on Microarchitecture (MICRO)}, 2020, pp. 622--636.

\bibitem{gamma}
S.-C. Kao and T.~Krishna, ``Gamma: Automating the hw mapping of dnn models on accelerators via genetic algorithm,'' in \emph{2020 IEEE/ACM International Conference On Computer Aided Design (ICCAD)}, 2020, pp. 1--9.

\bibitem{digamma}
S.-C. Kao, M.~Pellauer, A.~Parashar, and T.~Krishna, ``Digamma: Domain-aware genetic algorithm for hw-mapping co-optimization for dnn accelerators,'' in \emph{2022 Design, Automation \& Test in Europe Conference \& Exhibition (DATE)}, 2022, pp. 232--237.

\bibitem{Xiao2021HASCOTA}
Q.~Xiao, S.~Zheng, B.~Wu, P.~Xu, X.~Qian, and Y.~Liang, ``Hasco: Towards agile hardware and software co-design for tensor computation,'' \emph{2021 ACM/IEEE 48th Annual International Symposium on Computer Architecture (ISCA)}, pp. 1055--1068, 2021.

\bibitem{feng2023gandse}
L.~Feng, W.~Liu, C.~Guo, K.~Tang, C.~Zhuo, and Z.~Wang, ``Gandse: Generative adversarial network-based design space exploration for neural network accelerator design,'' \emph{ACM Transactions on Design Automation of Electronic Systems}, vol.~28, no.~3, pp. 1--20, 2023.

\bibitem{scalesim}
A.~Samajdar, J.~M. Joseph, Y.~Zhu, P.~Whatmough, M.~Mattina, and T.~Krishna, ``A systematic methodology for characterizing scalability of dnn accelerators using scale-sim,'' in \emph{2020 IEEE International Symposium on Performance Analysis of Systems and Software (ISPASS)}, 2020, pp. 58--68.

\bibitem{ramachandran2023ntrans}
A.~Ramachandran, A.~Dhiman, B.~S. Vandrotti, and J.~Kim, ``Ntrans-net: A multi-scale neutrosophic-uncertainty guided transformer network for indoor depth completion,'' in \emph{2023 IEEE International Conference on Image Processing (ICIP)}.\hskip 1em plus 0.5em minus 0.4em\relax IEEE, 2023, pp. 905--909.

\bibitem{maestro}
H.~Kwon, P.~Chatarasi, V.~Sarkar, T.~Krishna, M.~Pellauer, and A.~Parashar, ``{MAESTRO:} {A} data-centric approach to understand reuse, performance, and hardware cost of {DNN} mappings,'' \emph{{IEEE} Micro}, vol.~40, no.~3, pp. 20--29, 2020.

\bibitem{samajdar2020systematic}
A.~Samajdar, J.~M. Joseph, Y.~Zhu, P.~Whatmough, M.~Mattina, and T.~Krishna, ``A systematic methodology for characterizing scalability of dnn accelerators using scale-sim,'' in \emph{2020 IEEE International Symposium on Performance Analysis of Systems and Software (ISPASS)}.\hskip 1em plus 0.5em minus 0.4em\relax IEEE, 2020, pp. 58--68.

\bibitem{fu2022contrastive}
Y.~Fu, Q.~Yu, M.~Li, X.~Ouyang, V.~Chandra, and Y.~Lin, ``Contrastive quant: quantization makes stronger contrastive learning,'' in \emph{Proceedings of the 59th ACM/IEEE Design Automation Conference}, 2022, pp. 205--210.

\bibitem{cao2022synergistic}
Y.-H. Cao, P.~Sun, Y.~Huang, J.~Wu, and S.~Zhou, ``Synergistic self-supervised and quantization learning,'' in \emph{European Conference on Computer Vision}.\hskip 1em plus 0.5em minus 0.4em\relax Springer, 2022, pp. 587--604.

\bibitem{frumkin2023jumping}
N.~Frumkin, D.~Gope, and D.~Marculescu, ``Jumping through local minima: Quantization in the loss landscape of vision transformers,'' in \emph{Proceedings of the IEEE/CVF International Conference on Computer Vision}, 2023, pp. 16\,978--16\,988.

\bibitem{ramachandran2024clamp}
A.~Ramachandran, S.~Kundu, and T.~Krishna, ``Clamp-vit: Contrastive data-free learning for adaptive post-training quantization of vits,'' \emph{arXiv preprint arXiv:2407.05266}, 2024.

\bibitem{wang2023positive}
J.~Wang, J.~Li, W.~Li, L.~Xuan, T.~Zhang, and W.~Wang, ``Positive--negative equal contrastive loss for semantic segmentation,'' \emph{Neurocomputing}, vol. 535, pp. 13--24, 2023.

\bibitem{li2022targeted}
T.~Li, P.~Cao, Y.~Yuan, L.~Fan, Y.~Yang, R.~S. Feris, P.~Indyk, and D.~Katabi, ``Targeted supervised contrastive learning for long-tailed recognition,'' in \emph{Proceedings of the IEEE/CVF conference on computer vision and pattern recognition}, 2022, pp. 6918--6928.

\bibitem{roelofs2019meta}
R.~Roelofs, V.~Shankar, B.~Recht, S.~Fridovich-Keil, M.~Hardt, J.~Miller, and L.~Schmidt, ``A meta-analysis of overfitting in machine learning,'' \emph{Advances in Neural Information Processing Systems}, vol.~32, 2019.

\bibitem{vaswani2017attention}
A.~Vaswani, ``Attention is all you need,'' \emph{Advances in Neural Information Processing Systems}, 2017.

\bibitem{gomez2018automatic}
R.~G{\'o}mez-Bombarelli, J.~N. Wei, D.~Duvenaud, J.~M. Hern{\'a}ndez-Lobato, B.~S{\'a}nchez-Lengeling, D.~Sheberla, J.~Aguilera-Iparraguirre, T.~D. Hirzel, R.~P. Adams, and A.~Aspuru-Guzik, ``Automatic chemical design using a data-driven continuous representation of molecules,'' \emph{ACS central science}, vol.~4, no.~2, pp. 268--276, 2018.

\bibitem{miao2023occdepth}
R.~Miao, W.~Liu, M.~Chen, Z.~Gong, W.~Xu, C.~Hu, and S.~Zhou, ``Occdepth: A depth-aware method for 3d semantic scene completion,'' \emph{arXiv preprint arXiv:2302.13540}, 2023.

\bibitem{li2020generalized}
X.~Li, W.~Wang, L.~Wu, S.~Chen, X.~Hu, J.~Li, J.~Tang, and J.~Yang, ``Generalized focal loss: Learning qualified and distributed bounding boxes for dense object detection,'' \emph{Advances in Neural Information Processing Systems}, vol.~33, pp. 21\,002--21\,012, 2020.

\bibitem{resnet50}
K.~He, X.~Zhang, S.~Ren, and J.~Sun, ``Deep residual learning for image recognition,'' in \emph{Proceedings of the IEEE conference on computer vision and pattern recognition}, 2016, pp. 770--778.

\bibitem{touvron2023llama}
H.~Touvron, L.~Martin, K.~Stone, P.~Albert, A.~Almahairi, Y.~Babaei, N.~Bashlykov, S.~Batra, P.~Bhargava, S.~Bhosale \emph{et~al.}, ``Llama 2: Open foundation and fine-tuned chat models,'' \emph{arXiv preprint arXiv:2307.09288}, 2023.

\bibitem{dubey2024llama}
A.~Dubey, A.~Jauhri, A.~Pandey, A.~Kadian, A.~Al-Dahle, A.~Letman, A.~Mathur, A.~Schelten, A.~Yang, A.~Fan \emph{et~al.}, ``The llama 3 herd of models,'' \emph{arXiv preprint arXiv:2407.21783}, 2024.

\bibitem{mindmapping}
\BIBentryALTinterwordspacing
K.~Hegde, P.-A. Tsai, S.~Huang, V.~Chandra, A.~Parashar, and C.~W. Fletcher, ``Mind mappings: Enabling efficient algorithm-accelerator mapping space search,'' in \emph{Proceedings of the 26th ACM International Conference on Architectural Support for Programming Languages and Operating Systems}, ser. ASPLOS '21.\hskip 1em plus 0.5em minus 0.4em\relax New York, NY, USA: Association for Computing Machinery, 2021, p. 943–958. [Online]. Available: \url{https://doi.org/10.1145/3445814.3446762}
\BIBentrySTDinterwordspacing

\end{thebibliography}
